\documentclass[journal]{IEEEtran}
\usepackage{graphicx}
\usepackage{subfig}
\usepackage{amsmath}
\usepackage{url}
\usepackage{paralist}
\usepackage{subfig}
\usepackage{makecell}
\usepackage{amsfonts}
\usepackage{multirow}
\usepackage{algorithm}
\usepackage{algorithmic}

\ifCLASSINFOpdf
\else
\fi
\begin{document}
%

\title{A Deep Bag-of-Features Model for Music Auto-Tagging}

%
%
%

\author{Juhan Nam,~\IEEEmembership{Member,~IEEE,}
        Jorge Herrera, 
        and Kyogu Lee,~\IEEEmembership{Senior Member,~IEEE} \\ 
\thanks{J. Nam, J. Herrera and K. Lee are with Korea Advanced Institute of Science and Technology, South Korea,  Stanford University, CA, USA and Seoul National University, South Korea, respectively.}
}

\maketitle

\begin{abstract}
Feature learning and deep learning have drawn great attention in recent years as a way of transforming input data into more effective representations using learning algorithms. Such interest has grown in the area of music information retrieval (MIR) as well, particularly in music audio classification tasks such as auto-tagging. In this paper, we present a two-stage learning model to effectively predict multiple labels from music audio. The first stage learns to project local spectral patterns of an audio track onto a high-dimensional sparse space in an unsupervised manner and summarizes the audio track as a bag-of-features. The second stage successively performs the unsupervised learning on the bag-of-features in a layer-by-layer manner to initialize a deep neural network and finally fine-tunes it with the tag labels. Through the experiment, we rigorously examine training choices and tuning parameters, and show that the model achieves high performance on Magnatagatune, a popularly used dataset in music auto-tagging.    

 
\end{abstract}


%
\IEEEpeerreviewmaketitle

\section{Introduction}

In the recent past music has become ubiquitous as digital data. The scale of music collections that are readily accessible via online music services surpassed thirty million tracks\footnote{http://press.spotify.com/us/information/, accessed in Jan 23, 2015}. The type of music content has been also diversified as social media services allow people to easily share their own original music, cover songs or other media sources. These significant changes in the music industry have prompted new strategies for delivering music content, for example, searching a large volume of songs with different query methods (e.g., text, humming or audio example) or recommending a playlist based on user preferences. A successful approach to these needs is using meta data, for example, finding similar songs based on analysis by music experts or collaborative filtering based on user data. However, the analysis by experts is costly and limited, given the large scale of available music tracks. User data are intrinsically biased by the popularity of songs or artists. As a way of making up for these limitations of meta data, the audio content itself have been exploited, i.e., by training a system to predict high-level information from the music audio files. This content-based approach has been actively explored in the area of music information retrieval (MIR). They are usually formed as an audio classification task that predicts a single label given categories (e.g. genre or emotion) or multiple labels in  various aspects of music. The latter is often referred to as music annotation and retrieval, or simply called  \emph{music auto-tagging}. 

These audio classification tasks are generally implemented through two steps; feature extraction and supervised learning. While the supervised learning step is usually handled by commonly used classifiers such as Gaussian mixture model (GMM) and support vector machines (SVM), the feature extraction step has been extensively studied based on domain knowledge. For example, Tzanetakis and Cook in their seminal work on music genre classification presented comprehensive signal processing techniques to extract audio features that represents timbral texture, rhythmic content and pitch content of music \cite{Tzanetakis:02}. Specifically, they include low-level spectral summaries (e.g. centroid and roll-off), zero-crossings and mel-frequency cepstral coefficients (MFCC), a wavelet transform-based beat histogram and pitch/chroma histogram. McKinney and Breebaart suggested perceptual audio features based on psychoacoustic models, including estimates of roughness, loudness and sharpness, and auditory filterbank temporal envelopes \cite{McKinney:03}. Similarly, a number of audio features have been proposed with different choices of time-frequency representations, psychoacoustic models and other signal processing techniques. Some of distinct audio features introduced in music classification include octave-based spectral contrast \cite{Jiang:02}, Daubechies wavelet coefficient histogram \cite{Li:03}, and auditory temporal modulation \cite{Panagakis:10}. 

A common aspect of these audio features is that they are \emph{hand-engineered}. In other words, individual computation steps to extract the features from audio signals are manually designed based on signal processing and/or acoustic knowledge. Although this hand-engineering approach has been successful to some degree, it has limitations in that, by nature, it may require numerous trial-and-error in the process of fine-tuning the computation steps. For this reason, many of previous work rather combine existing audio features, for example, by concatenating MFCC and other spectral features \cite{Bergstra:06, T.Bertin-Mahieux:08, Chang:10}. However, they are usually heuristically chosen so that the combination can be redundant or still insufficient to explain music. Feature selection is a solution to finding an optimal combination but this is another challenge \cite{Silla:08}.

Recently there have been increasing interest in finding feature representations using data-driven learning algorithms, as an alternative to the hand-engineering approach. Inspired by research in computational neuroscience \cite{B.Olshausen:96, M.Lewicki:02},  the machine learning community has developed a variety of learning algorithms that discover underlying structures of image or audio, and utilized them to represent features. This approach made it possible to overcome the limitations of the hand-engineering approach by learning manifold patterns automatically from data. This learning-based approach is broadly referred to as \emph{feature learning} or \emph{representation learning} \cite{Y.Bengio:13}. In particular, hierarchical representation learning based on deep neural network (DNN) or convolutional neural network (CNN), called \emph{deep learning}, achieved a remarkable series of successes in challenging machine recognition tasks, such as speech recognition \cite{G.Hinton:12} and image classification \cite{A.Krizhevsky:12}. The overview and recent work are reviewed in \cite{Y.Bengio:13,Y.Bengio:09}. The learning-based approach has gained great interest in the MIR community as well. Leveraging advances in the machine learning community, MIR researchers have investigated better ways of representing audio features and furthermore envisioned the approach as a general framework to build hierarchical music feature representations \cite{E.Humphrey:12}. In particular, the efforts have been made most actively for music genre classification or music auto-tagging. Using either unsupervised feature learning or deep learning, they have shown improved performance in the tasks.





In this paper, we present a two-stage learning model as an extension of our previous work \cite{J.Nam:12}. The first stage learns local features from multiple frames of audio spectra using sparse restricted Boltzmann machine (RBM) as before. However, we add an onset detection module to select temporally-aligned frames in training data. This is intended to decrease the variation in input space against random sampling. We show that this helps improving performance given the same condition. The second stage continues the bottom-up unsupervised learning by applying RBMs (but without sparsity) to the bag-of-features in a layer-by-layer manner. We use the RBM parameters to initialize a DNN and finally fine-tunes the network with the labels. We show that this pretraining improves the performance as observed in image classification or speech recognition tasks. 





The remainder of this paper is organized as follows. In Section \ref{related_work}, we overview related work. In Section \ref{learning_model}, we describe the bag-of-features model. In Section \ref{experiments}, we introduce datasets, evaluation metrics and experiment settings. In Section \ref{evaluation}, we investigate the the evaluation results and compare them to those of state-of-the-arts algorithms in music auto-tagging. Lastly, we conclude by providing a summary of the work in Section \ref{conclusion}. 

\section{Related Work} \label{related_work}

In this section, we review previous work that exploited feature learning and deep learning for music classification and music auto-tagging. They can be divided into two groups, depending on whether the learning algorithm is unsupervised or supervised.  

One group investigated unsupervised feature learning based on sparse representations, for example, using K-means \cite{J.Wulfing:12, J.Nam:12, S.Dieleman:13, A.Oord:14}, sparse coding \cite{R.Grosse:07, P.Manzagol:08, M.Henaff:11, J.Nam:12, Vaizman:14} and restricted Boltzmann machine (RBM) \cite{Schlueter:11, J.Nam:12}. The majority of them focused on capturing local structures of music data over one or multiple audio frames to learn high-dimensional single-layer features. They summarized the locally learned features as \emph{a bag-of-features} (also called \emph{a bag-of-frames}, for example, in \cite{Su:14}) and fed them into a separate classifier. The advantage of this single-layer feature learning is that it is quite simple to learn a large-size of feature bases and they generally provide good performance \cite{A.Coates:11b}. In addition, it is easy to handle the variable length of audio tracks as they usually represent song-level features with summary statistics of the locally learned features (i.e. temporal pooling). However, this single-layer approach is limited to learning local features only. Some works worked on dual or multiple layers to capture segment-level features \cite{Lee+etal09:convDBNAudio, C.Yeh:13}. Although they showed slight improvement by combining the local and segment-level features, learning hierarchical structures of music in an unsupervised way is highly challenging. 



\begin{figure*}[t]
  \centering
   \includegraphics[trim=5mm 0mm 0mm 0mm, scale=.7]{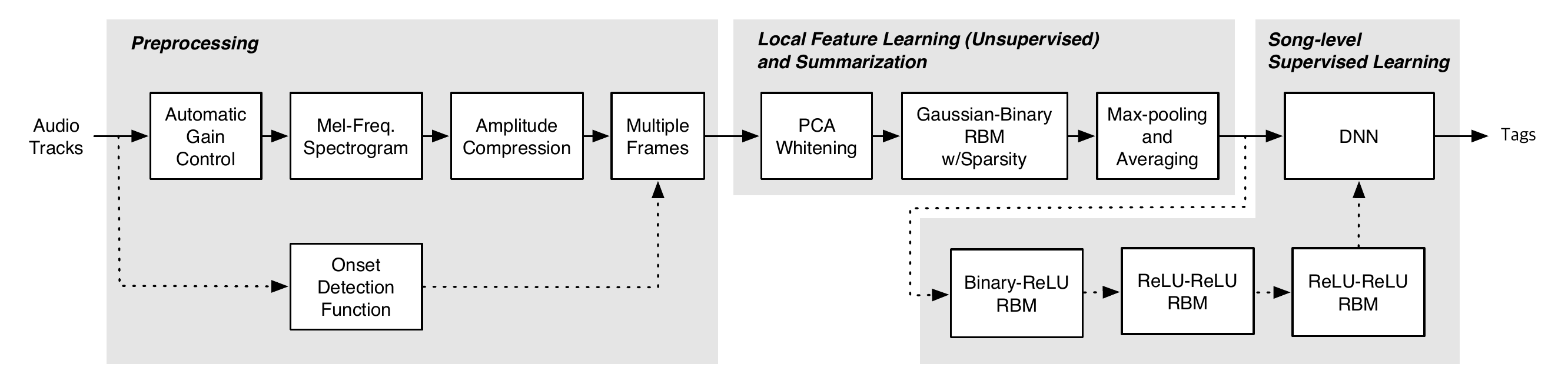}
  \caption{The proposed deep bag-of-features model for music auto-tagging. The dotted lines indicate that the processing is conducted only in the training phase.}
  \label{fig:pipeline}
\end{figure*}

The second group used supervised learning that directly maps audio and labels via multi-layered neural networks. One approach was mapping single frames of spectrogram \cite{P.Hamel:10, E.Schmidt:11, S.Sigtia:14} or summarized spectrogram \cite{E.Schmidt:12} to labels via DNNs, where some of them pretrain the networks with deep belief networks \cite{P.Hamel:10, E.Schmidt:11, E.Schmidt:12}. They used the hidden-unit activations of DNNs as local audio features. While this frame-level audio-to-label mapping is somewhat counter-intuitive, the supervised approach makes the learned features more discriminative for the given task, being directly comparable to hand-engineered features such as MFCC. The other approach in this group used CNNs where the convolution setting can take longer audio frames and the networks directly predict labels \cite{P.Hamel:11, P.Hamel:12, A.Oord:13, S.Dieleman:14}. CNNs has become the de-facto standard in image classification since the break-through in ImageNet challenge \cite{A.Krizhevsky:12}. As such, the CNN-based approach has shown great performance in music auto-tagging \cite{P.Hamel:12, S.Dieleman:14}. However, in order to achieve high performance with CNNs, the model needs to be trained with a large dataset along with a huge number of parameters. Otherwise, CNNs is not necessarily better than the bag-of-features approach  \cite{S.Dieleman:13, S.Dieleman:14}. 

Our approach is based on the bag-of-features in a single-layer unsupervised learning but extend it to a deep structure for song-level supervised learning. The idea behind this deep bag-of-features model is, while taking the simplicity and flexibility of the bag-of-features approach in unsupervised single-layer feature learning, improving the discriminative power using deep neural networks. Similar models were suggested using a different combination of algorithms, for example, K-means and multi-layer perceptrons (MLP) in \cite{S.Dieleman:13, A.Oord:14}. However, our proposed model performs unsupervised learning through all layers consistently using RBMs.  



\section{Learning Model} \label{learning_model}

Figure \ref{fig:pipeline} illustrates the overall data processing pipeline of our proposed model. In this section, we describe the individual processing blocks in details.  

 
\subsection{Preprocessing}

Musical signals are characterized well by note onsets and ensuing spectral patterns. We perform several steps of front-end processing to help learning algorithms effectively capture the features.  


\subsubsection{Automatic Gain Control}
Musical signals are highly dynamic in amplitude. Being inspired by the dynamic-range compression mechanism in human ears, we control the amplitude as a first step. We adopt time-frequency automatic gain control which adjusts the levels of sub-band signals separately \cite{D.Ellis:10}. Since we already showed the effectiveness in music auto-tagging \cite{J.Nam:12}, we use the automatic gain control as a default setting here.

\subsubsection{Mel-frequency Spectrogram}
We use mel-frequency spectrogram as a primary input representation. The mel-frequency spectrogram is computed by mapping 513 linear frequency bins from FFT to 128 mel-frequency bins. This mapping reduces input dimensionality sufficiently so as to take multiple frames as input data while preserving distinct patterns of spectrogram well.

\subsubsection{Amplitude Compression}
The mel-frequency spectrogram is additionally compressed with a log-scale function, $\log_{10}(1+C \cdot x)$, where $x$ is the mel-frequency spectrogram and $C$ controls the degree of compression \cite{M.Mueller:11}. 

\subsubsection{Onset Detection Function} 
\begin{figure} [t]
  \centering
   \includegraphics[trim=4mm 0mm 00mm 0mm, width=88mm, height=65mm]{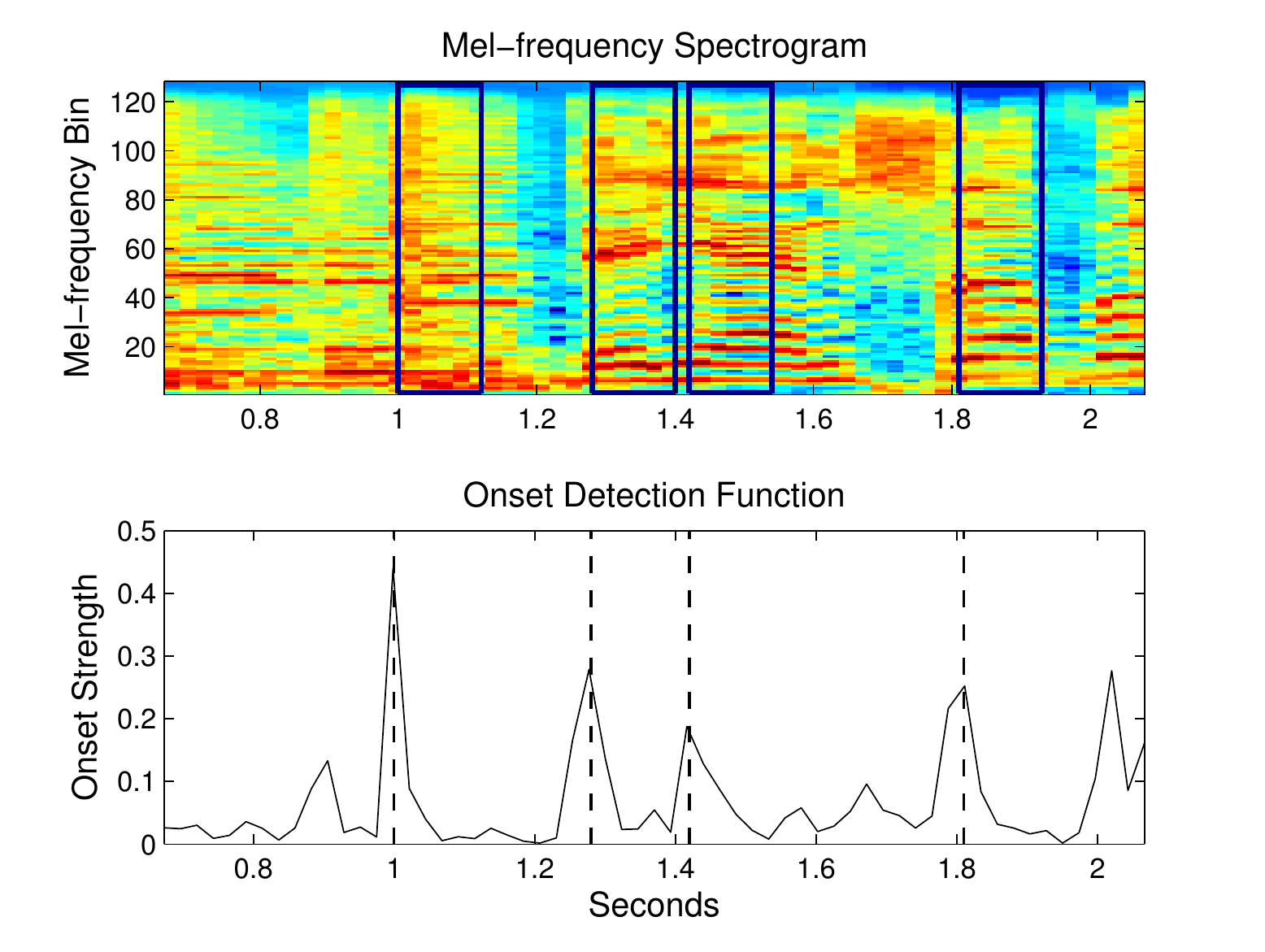}
  \caption{Onset-based sampling. This data sampling scheme takes multiple frames at the positions that the onset strength is high.}
  \label{fig:sample_selection}
\vspace{-3mm}  
\end{figure}

The local feature learning stage takes multiple frames as input data so that learning algorithms can capture spectro-temporal patterns, for example, sustaining or chirping harmonic overtones, or transient changes over time. We already showed that using multiple frames for feature learning improves the performance in music auto-tagging \cite{J.Nam:12}. We further develop this data selection scheme by considering where to take multiple frames on the time axis. In the previous work, we sampled multiple frames at random positions on the mel-frequency spectrogram without considering the characteristics of musical sounds. Therefore, given a single note, it could sample audio frames such that the note onset is located at arbitrary positions within the sampled frames or only sustain part of a note is taken. This may increase unnecessary variations or lose the chance of capturing important temporal dependency in the view of learning algorithm. In order to address this problem, we suggest sampling multiple frames based on the guidance of note onset. That is, we compute an onset detection function as a separate path and take a sample of multiple frames at the positions that the onset detection function has high values for a short segment. As illustrated in Figure \ref{fig:sample_selection}, local spectral structures of musical sounds tend to be more distinctive when the onset strength is high. Sampled audio frames this way are likely to be aligned to each other with regard to notes, which may encourage learning algorithms to learn features more effectively. We term this sampling scheme \emph{onset-based sampling} and will evaluate it in our experiment. The onset detection function is computed on a separate path by mapping the spectrogram on to 40 sub-bands and summing the half-wave rectified spectral flux over the sub-bands.

\subsection{Local Feature Learning and Summarization} \label{algorithm}

This stage first learns feature bases using the sampled data and learning algorithms. Then, it extracts the feature activations in a convolutional manner for each audio track and summarizes them as a bag-of-features using max-pooling and averaging.     

\subsubsection{PCA Whitening} 
PCA whitening is often used as a preprocessing step to remove pair-wise correlations (i.e. second-order dependence) or reduce the dimensionality before applying algorithms that capture higher-order dependencies \cite{Hyvarinen:09}. The PCA whitening matrix is computed by applying PCA to the sampled data and normalizing the output in the PCA space. Note that we locate PCA whitening as part of local feature learning in Figure \ref{fig:pipeline} because the whitening matrix is actually ``learned'' by the sampled data.

\vspace{1mm}
\subsubsection{Sparse Restricted Boltzmann Machine (RBM)}
Sparse RBM is the core algorithm that performs local feature learning in the bag-of-features model. In our previous work, we compared K-means, sparse coding and sparse RBM in terms of performance of music auto-tagging \cite{J.Nam:12}. Although there was not much difference, sparse RBM worked slightly better than others and the feed-forward computation for the hidden units in the RBM allows fast prediction in the testing phase. Thus, we focus only the sparse RBM here and more formally review the algorithm in the following paragraphs. 

Sparse RBM is a variation of RBM which is a bipartite undirected graphical model that consists of visible nodes $\textbf{v}$ and hidden nodes $\textbf{h}$ \cite{Lee+etal08:sparseDBN}. The visible nodes correspond to input vectors in a training set and the hidden nodes correspond to represented features. The basic form of RBM has binary units for both visible and hidden nodes, termed \emph{binary-binary} RBM. The joint probability of $\textbf{v}$ and $\textbf{h}$ is defined by an energy function $E(\textbf{v}, \textbf{h})$: 
 
\newcommand\minus{%
  \setbox0=\hbox{-}%
  \vcenter{%
    \hrule width\wd0 height \the\fontdimen8\textfont3%
  }%
}

\begin{equation}
p(\textbf{v}, \textbf{h}) = \frac{e^{-E(\textbf{v}, \textbf{h})}}{Z}
\label{eq:rbmeqn1}
\end{equation}

\begin{equation}
\label{eq:rbmeqn2}
 E(\textbf{v}, \textbf{h}) = - \left( \textbf{b}^{T}\textbf{v} + \textbf{c}^{T}\textbf{h} + \textbf{v}^{T}\textbf{W}\textbf{h} \right) 
\end{equation}
where $\textbf{b}$ and $\textbf{c}$ are bias terms, and $\textbf{W}$ is a weight matrix. The normalization factor $Z$ is called the partition function, which is obtained by summing all possible configurations of $\textbf{v}$ and $\textbf{h}$. For real-valued data such as spectrogram, Gaussian units are frequently used for the visible nodes. Then, the energy function in Equation \ref{eq:rbmeqn1} is modified to:
\begin{equation}
\label{eq:rbmeqn}
 E(\textbf{v}, \textbf{h}) = \textbf{v}^{T}\textbf{v} - \left( \textbf{b}^{T}\textbf{v} + \textbf{c}^{T}\textbf{h} + \textbf{v}^{T}\textbf{W}\textbf{h} \right) 
\end{equation}
where the additional quadratic term, $\textbf{v}^{T}\textbf{v}$ is associated with covariance between input units assuming that the Gaussian units have unit variances. This form is called \emph{Gaussian-binary} RBM \cite{Y.Bengio:07}.

The RBM has symmetric connections between the two layers but no connections within the hidden nodes or visible nodes. This conditional independence makes it easy to compute the conditional probability distributions, when nodes in either layer are observed: 
\begin{equation}
\label{eq:CondHGivenV}
p(h_j = 1| \textbf{v}) = \sigma(c_j + \sum_{i}W_{ij}v_i) \qquad\qquad 
\vspace{-2mm}
\end{equation}
\begin{equation}
\label{eq:CondVGivenH}
p(v_i | \textbf{h}) =  \mathcal{N}( b_i + \sum_{j}W_{ij}h_j, 1), \qquad\qquad 
\end{equation}
where $\sigma(x) = 1/(1+\exp(\minus x))$ is the logistic function and $\mathcal{N}(x)$ is the Gaussian distribution. These can be directly derived from Equation \ref{eq:rbmeqn1} and \ref{eq:rbmeqn2}.

The model parameters of RBM are estimated by taking derivative of the log-likelihood with regard to each parameter and then updating them using gradient descent. The update rules for weight matrix and bias terms are obtained from Equation \ref{eq:rbmeqn1}:  
\begin{equation}
W_{ij} \leftarrow W_{ij} + \epsilon(\left<v_{i}h_{j}\right>_{data} - \left<v_{i}h_{j}\right>_{model})
\label{eq:rbm_ml_w}
\end{equation}
\begin{equation}
b_{i} \leftarrow b_{i} + \epsilon(\left<v_{i}\right>_{data} - \left<v_{i}\right>_{model})
\label{eq:rbm_ml_b}
\end{equation}
\begin{equation}
c_{j} \leftarrow c_{j} + \epsilon(\left<h_{j}\right>_{data} - \left<h_{j}\right>_{model})
\label{eq:rbm_ml_c}
\end{equation}
where $\epsilon$ is the learning rate and the angle brackets denote expectation with respect to the distributions from the training data and the model. While $\left<v_{i}h_{j}\right>_{data}$ can be easily obtained, exact computation of $\left<v_{i}h_{j}\right>_{model}$ is intractable. In practice, the learning rules in Equation \ref{eq:rbm_ml_w} converges well only with a single iteration of block Gibbs sampling when it starts by setting the states of the visible units to the training data. This approximation is called the \emph{contrastive-divergence} \cite{G.Hinton:06}. 

This parameter estimation is solely based on maximum likelihood and so it is prone to overfitting to the training data. As a way of improving generalization to new data \cite{G.Hinton:10}, the maximum likelihood estimation is penalized with additional terms called weight-decay. The typical choice is $L_{2}$ norm, which is half of the sum of the squared weights. Taking the derivative, the weight update rule in Equation \ref{eq:rbm_ml_w} is modified to:
\begin{equation}
 W_{ij} \leftarrow  W_{ij} + \epsilon (\left<v_{i}h_{j}\right>_{data}-\left<v_{i}h_{j}\right>_{model}-\mu W_{ij}) 
\label{eq:rbm_ml_w_weight_decay}   
\end{equation}
where $\mu$ is called \emph{weight-cost} and controls the strength of the weight-decay. 

Sparse RBM is trained with an additional constraint on the update rules, which we call sparsity. We impose the sparsity on hidden units of a Gaussian-binary RBM based on the technique in \cite{Lee+etal08:sparseDBN}. The idea is to add a penalty term that minimizes a deviation of the mean activation of hidden units from a target sparsity level. Instead of directly applying the gradient descent to that, they exploit the contrastive-divergence update rule and so simply added it to the update rule of bias term $c_{j}$. This controls the hidden-unit activations as a shift term of the sigmoid function in Equation \ref{eq:CondHGivenV}. As a result, the bias update rule in Equation \ref{eq:rbm_ml_c} is modified to: 
\begin{equation} 
c_{j} \leftarrow c_{j} + \epsilon (\left<h_{j}\right>_{data} - \left<h_{j}\right>_{model}) + \lambda \sum_j (\rho - \frac{1}{m} (\sum_{k=1}^m \left<h_j|\textbf{v}^k\right>))^2, 
\label{sparsity_constraint}
\end{equation}
where $\{\textbf{v}^1,...,\textbf{v}^m\}$ is the training set,  $\rho$ determines the target sparsity of the hidden-unit activations and $\lambda$ controls the strength. 

\vspace{1mm}
\subsubsection{Max-Pooling and Averaging}

Once we train a sparse RBM from the sampled data, we fix the learned parameters and extract the hidden-unit activations in a convolutional manner for an audio track. Following our previous work, we summarize the local features via max-pooling and averaging. Max-pooling has proved to be an effective choice to summarize local features \cite{P.Hamel:11, S.Dieleman:13}. It works as a form of temporal masking because it discards small activations around high peaks. We further summarize the max-pooled feature activations with averaging. This produces a bag-of-features that represents a histogram of dominant local feature activations.




\begin{figure} [t]
  \centering  
  \includegraphics[trim=7mm 0mm 0mm 0mm, scale=0.7]{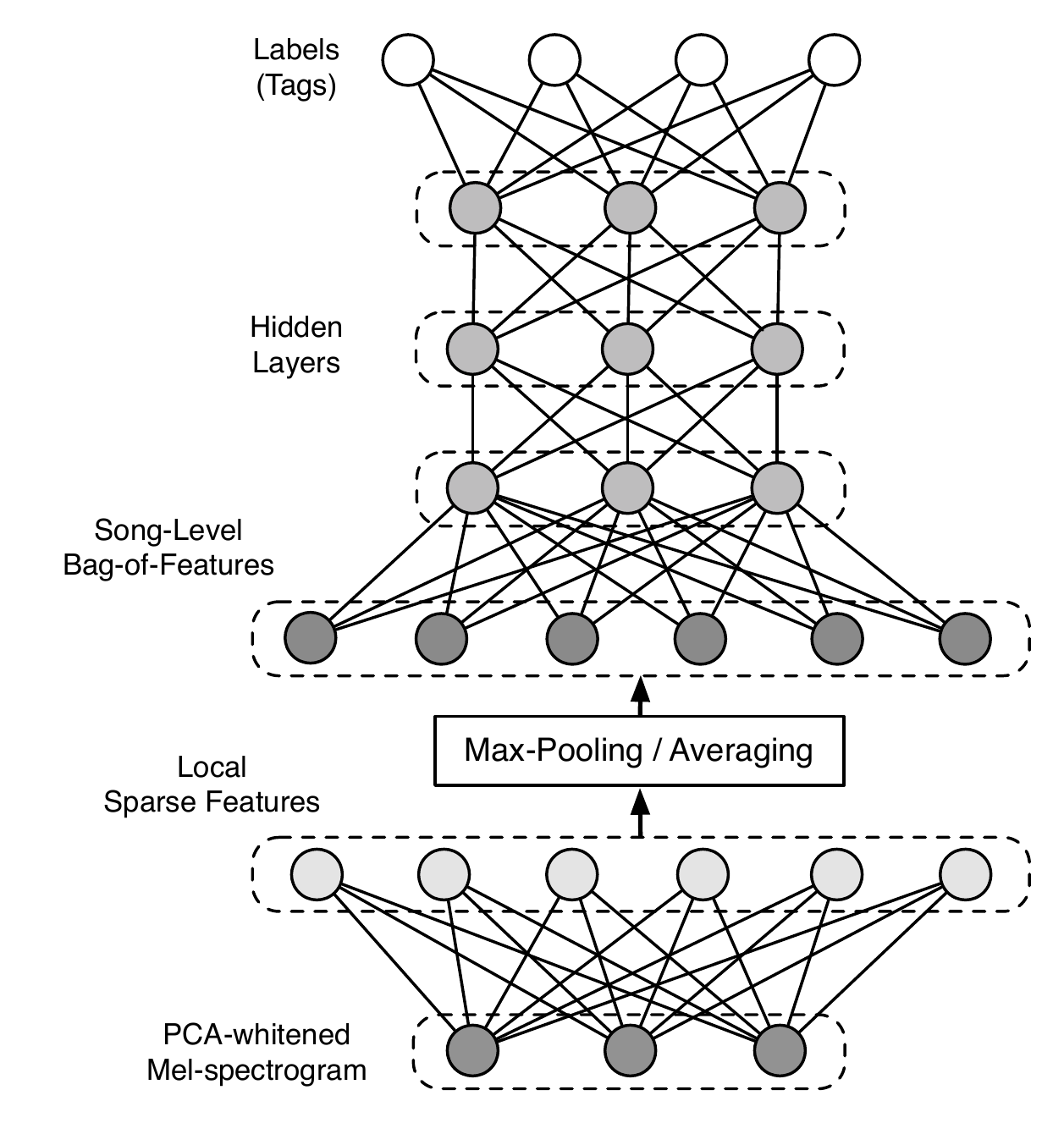}
\vspace{3mm}
  \caption{The architecture of bag-of-features model. The three fully connected layers are first pretrained with song-level bag-of-features data using stacked RBMs with ReLU and then fine-tuned with labels.}
  \label{fig:architecture_of_BOF_model}
\end{figure}

\subsection{Song-Level Learning}
This stage performs supervised learning to predict tags from the bag-of-features. Using a deep neural network (DNN), we build a \emph{deep bag-of-features} representation that maps the complex relations between the summarized acoustic features and semantic labels. We configure the DNN to have up to three hidden layers and rectified linear units (ReLUs) for the nonlinear function as shown in Figure \ref{fig:architecture_of_BOF_model}. The ReLUs have proved to be highly effective in the DNN training when used with the dropout regularization \cite{M.Zeiler:13, G.Dahl:13, S.Sigtia:14}, and also much faster than other nonlinear functions such as sigmoid in computation. We first pretrain the DNN by a stack of RBMs and then fine-tune it using tag labels. The output layer works as multiple independent binary classifiers. Each output unit corresponds to a tag label and predicts whether the audio track is labeled with it or not. 

\vspace{1mm}

\subsubsection{Pretraining}
Pretraining is an unsupervised approach to better initialize the DNN \cite{G.Hinton:06}. Although recent advances have shown that pretraining is not necessary when the number of labeled training samples is sufficient \cite{A.Krizhevsky:12, D.Yu:15}, we conduct the pretraining to verify the necessity in our experiment setting. 

We perform the pretraining by greedy layer-wise learning of RBMs with ReLUs to make learned parameters compatiable with the nonlinearity in the DNN. The ReLUs in the RBMs can be viewed as the sum of an infinite number of binary units that share weights and have shifted versions of the same bias \cite{V.Nair:10}. This can be approximated to a single unit with the $\max(0,x)$ nonlinearity. Furthermore, Gibbs sampling for the ReLUs during the training can be performed by taking samples from $\max(0,x + \mathcal{N}(0,\sigma(x)))$ where $\mathcal{N}(0,\sigma(x))$ is Gaussian noise with zero mean and variance $\sigma(x)$ \cite{V.Nair:10}. We use the ReLU for both visible and hidden nodes of the stacked RBMs. However, for the bottom RBM that takes the bag-of-features as input data, we use binary units for visible nodes and ReLU for hidden notes to make them compatible with the scale of the bag-of-features. 

    


\vspace{1mm}
\subsubsection{Fine-tuning}

After initializing the DNN with the weight and bias learned from the RBMs, we fine-tune them with tag labels using the error back-propagation. We predict the output by adding the output layer (i.e. weight and bias) to the last hidden layer and taking the sigmoid function to define the error as cross-entropy between the prediction $h_{\theta,j}(x_i)$ and ground truth $y_{ij} \in \left\{0,1 \right\} $ for bag-of-features $i$ and tag $j$:

\begin{equation} 
J(\theta) = \sum_i \sum_j y_{ij}\log(h_{\theta,j}(x_i)) + (1-y_{ij})(1-\log(h_{\theta,j}(x_i)))
\label{cost_function}
\end{equation}
We update a set of parameters $\theta $ using AdaDelta. The method requires no manual tuning for the learning rate and is robust to noisy gradient information and variations in model architecture \cite{M.Zeiler:12}. In addition, we use dropout, a powerful technique that improves the generalization error of large neural networks by setting zeros to hidden and input layers randomly \cite{N.Srivastava:14}. We find AdaDelta and dropout essential to achieve good performance.   


\section{Experiments} \label{experiments} 
In the section, we introduce the dataset and evaluation metrics used in our experiments. Also, we describe experiment settings for the proposed model.

\subsection{Datasets} 
We use the Magnatagatune dataset, which contains 29-second MP3 files with annotations collected from an online game \cite{E.Law:09}. The dataset is the MIREX 2009 version used in \cite{P.Hamel:11, P.Hamel:12}. It is split into 14660, 1629 and 6499 clips, respectively for training, validation and test, following the prior work. The clips are annotated with a set of 160 tags. 
 
\subsection{Evaluation Metrics}
Following the evaluation metrics in \cite{P.Hamel:11,P.Hamel:12}, we use the area under the receiver operating characteristic curve over tag (AUC-T or shortly AUC), the same measure over clip (AUC-C) and top-K precision where K is 3, 6, 9, 12 and 15. 


\subsection{Preprocessing Parameters}
We first convert the MP3 files to the WAV format and resample them to 22.05kHz. We then compute their spectrogram with a 46ms Hann window and 50\% overlap, on which the time-frequency automatic gain control using the technique in \cite{D.Ellis:10} is applied. This equalizes the spectrogram using spectral envelopes computed over 10 sub-bands. We convert the equalized spectrogram to mel-frequency spectrogram with 128 bins and finally compress the magnitude by fixing the strength $C$ to 10.

\subsection{Experiment Settings}

\subsubsection{Local Feature Learning and Summarization}
The first step in this stage is to train PCA (for whitening) and sparse RBM. Each training sample is a spectral block comprised of multiple consecutive frames from the mel-frequency spectrogram. We gather training data (total 200,000 samples) by taking one spectral block every second at a random position or using the onset detection function. The number of frames in the spectral block varies from 2, 4, 6, 8 to 10 and we evaluate them separately. We obtain the PCA whitening matrix retaining 90\% of the variance to reduce the dimensionality and then train the sparse RBM with a learning rate of 0.03, a hidden-layer size of 1024 and different values of target sparsity $\rho$ from 0.007, 0.01, 0.02 to 0.03. Once we learn the PCA whitening matrix and RBM weight, we extract hidden-unit activations from an audio track in a convolutional manner and summarize them into a bag-of-features with max-pooling over segments with 0.25, 0.5, 1, 2 and 4 seconds. 

Since this stage creates a large number of possibilities in obtaining a bag-of-features, we reduce the number of adjustable parameters before proceeding with song-level supervised learning. Among others, we fix the number of frames in the spectral block and data sampling scheme, which are related to collecting the sample data. We find a reasonable setting for them using a simple linear classifier that minimizes the same cross-entropy in Equation \ref{cost_function} (i.e. logistic regression).

\begin{figure} [t]
  \centering  
  \includegraphics[trim=7mm 0mm 0mm 0mm, width=95mm, height=58mm]{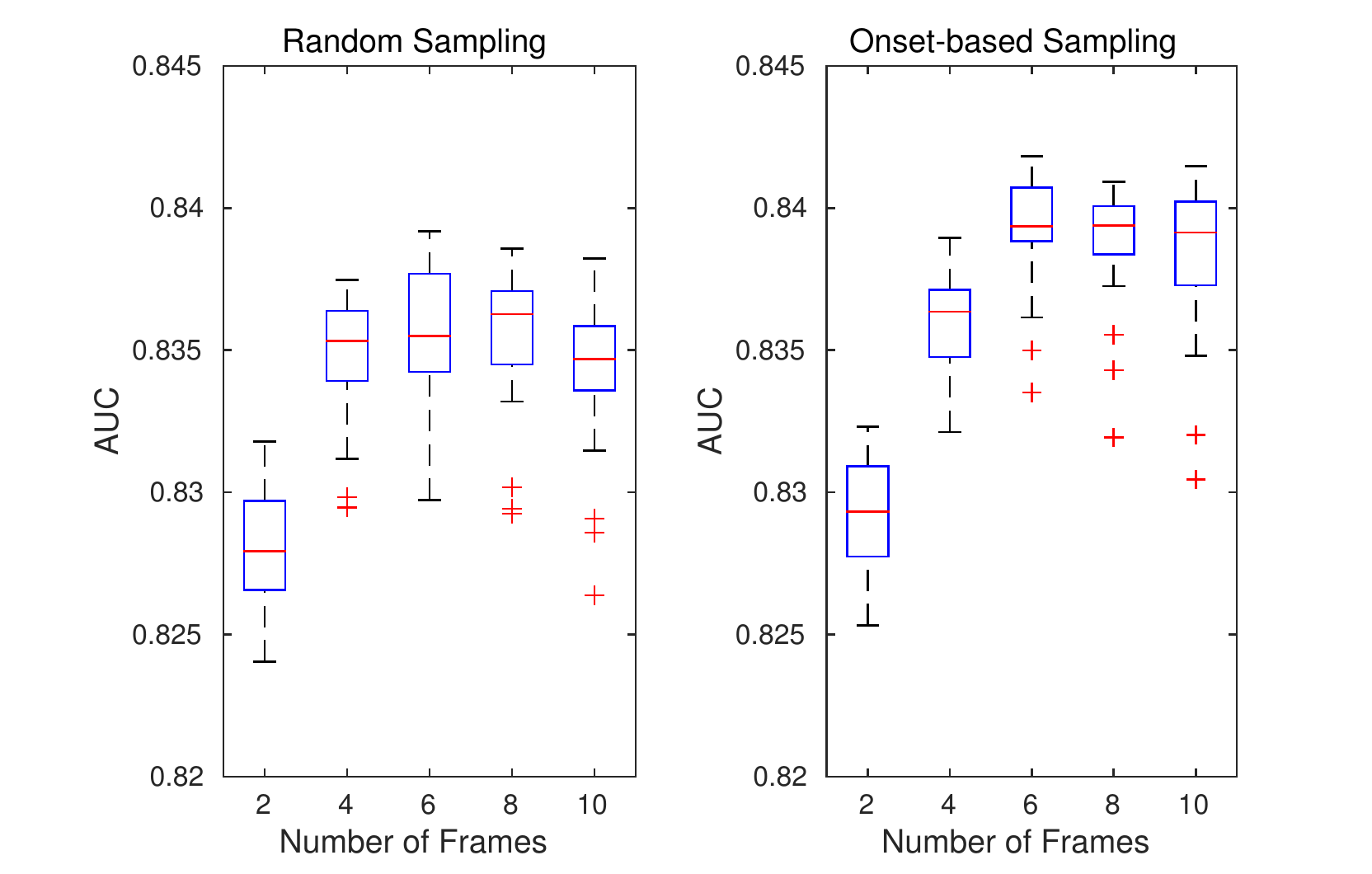}
  \caption{Results for number of frames in the input spectral block and data sampling scheme. Each box contains the statistics of AUC for different sparsity and max-pooling sizes.}
  \label{fig:effect_of_frame_size_and_onset_preferred_sampling}
\end{figure}

\begin{figure}[t!]
\centering
   \includegraphics[trim=8mm -2mm 0mm 0mm, scale=0.58]{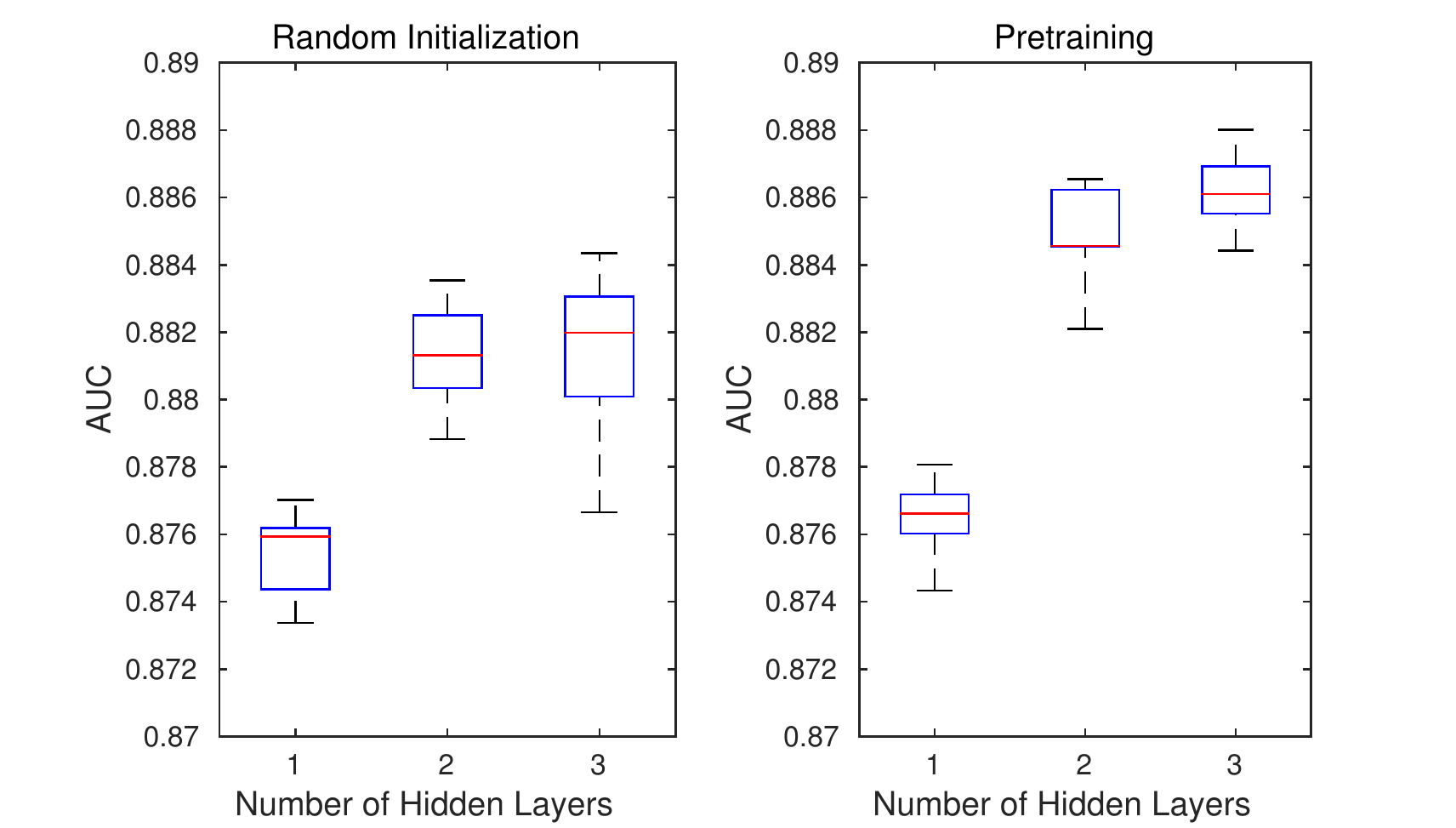}
 \caption{Results for different number of hidden layers. Each boxplot contains the statistics of AUC for different target sparsity and max-pooling sizes in the bag-of-features.}
  \label{fig:aroc_number_of_hidden_layers}
\end{figure}

\subsubsection{Song-Level Supervised Learning} \label{tuninng_params_song_level}
We first pretrain the DNN with RBMs and then fine-tune the networks. We fix the hidden-layer size to 512 and adjust the number of hidden layers from 1 to 3 to verify the effect of larger networks. In training ReLU-ReLU RBMs, we set the learning rate to a small value (0.003) in order to avoid unstable dynamics in the weight updates \cite{G.Hinton:10}. We also adjust the weight-cost in training RBMs from 0.001, 0.01 to 0.1, separately for each hidden layer. We fine-tune the pretrained networks, using Deepmat, a Matlab library for deep learning\footnote{\url{https://github.com/kyunghyuncho/deepmat}}. This library includes an implementation of AdaDelta and dropout, and supports GPU processing. In order to validate the proposed model, we compare it to DNNs with random initialization and also the same model but with ReLU units for the visible layer of the bottom RBM\footnote{Our experiment code is available at \url{https://github.com/juhannam/deepbof}}

\section{Results} \label{evaluation}
In this section, we examine training choices and tuning parameters in the experiments, and finally compare them to state-of-the-art results.  





\subsection{Onset Detection Function and Number of Frames} 
We compare random sampling with onset-based sampling in the context of finding an optimal number of frames in local feature learning. In order to prevent the experiment from being too exhausting, we chose logistic regression as a classifier instead of the DNN. Figure \ref{fig:effect_of_frame_size_and_onset_preferred_sampling} shows the evaluation results. In random sampling, the AUC increases up to 6 frames and then slowly decays. A similar trend is shown in onset-based sampling. However, the AUC saturates in a higher level, indicating that onset-based sampling is more effective for the local feature learning. In the following experiments, we fix the number of frames to 8 as it provides the highest AUC in terms of median.

\subsection{Pretraining by ReLU RBMs}
Figure \ref{fig:aroc_number_of_hidden_layers} shows the evaluation results for different numbers of hidden layers when the DNN is randomly initialized or pretrained with ReLU RBMs. When the networks has a single hidden layer, there is no significant difference in AUC level. As the number of hidden layers increases in the DNN, however, pretrained networks  apparently outperform randomly initialized networks. This result is interesting, when recalling recent observations that pretraining is not necessary when the number of labeled training samples is sufficient. Thus, the result may indicate that the size of labeled data is not large enough in our experiment. However, we need to note that the auto-tagging task is formed as a multiple binary classification problem, which is different from choosing one label exclusively, and furthermore the levels of abstraction in the tag labels are not homogenous (e.g. including mood and instruments). In addition, there is some recent work that pretraining is still useful \cite{G.Dahl:13}.




\begin{figure}[t!]
\centering
   \includegraphics[trim=5mm -2mm 0mm 0mm, scale=0.58]{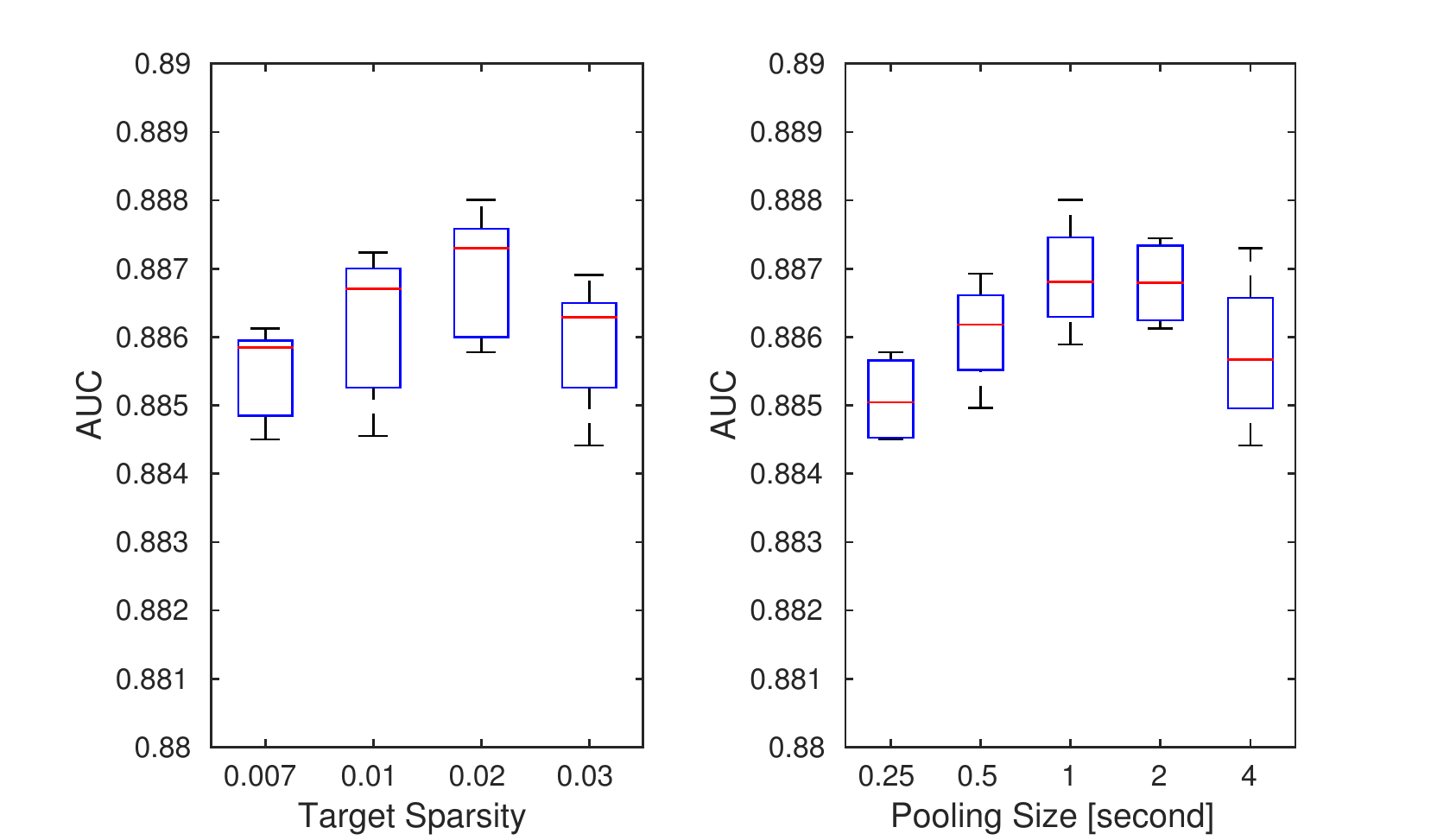}
 \caption{Results for different target sparsity and max-pooling sizes in the bag-of-features when we use a pretrained DNN with three hidden layers.}
  \label{fig:aroc_sparsity_pooling}
\end{figure}

\subsection{Sparsity and Max-pooling}
Figure \ref{fig:aroc_sparsity_pooling} shows the evaluation results for different target sparsity and max-pooling sizes in the bag-of-features when we use a pretrained DNN with three hidden layers. The best results are achieved when target sparsity is 0.02 and max-pooling size is 1 or 2 second. Compared to our previous work \cite{J.Nam:12}, the optimal target sparsity has not changed whereas the optimal max-pooling size is significantly reduced. Considering we used 30 second segments in the Maganatagatune dataset against the full audio tracks in the CAL500 datasets (typically 3-4 minute long), the optimal max-pooling size seems to be proportional to the length of audio tracks.     

\subsection{Weight-Cost}
\begin{figure}[t!]
\centering
   \includegraphics[trim= 2mm -5mm 0mm 0mm, scale=0.63]{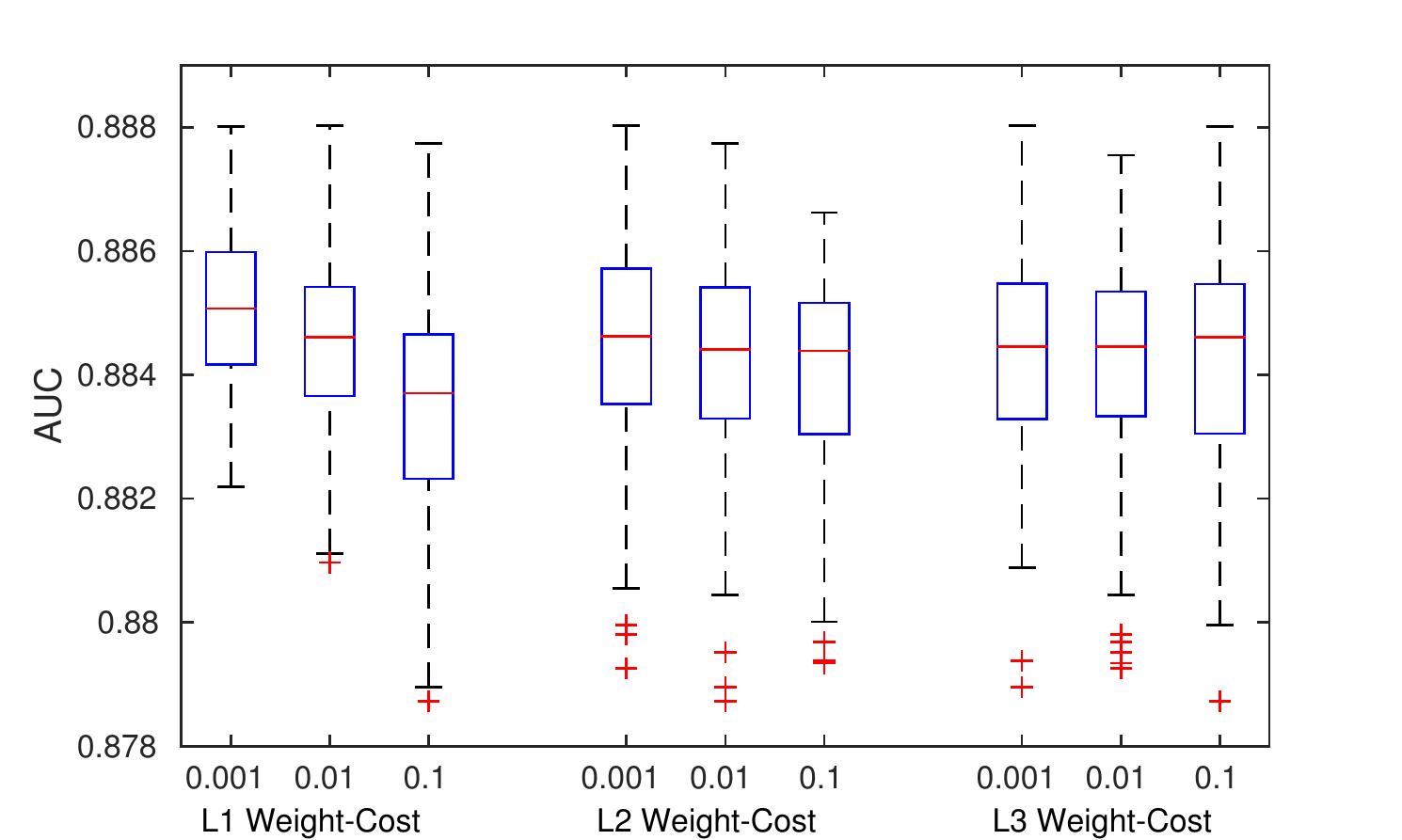}
 \caption{Results for a fixed weight-cost at each layer. Each boxplot contains the statistics of AUC for all weight-cost combinations in three hidden layers given the fixed weight-cost. }
  \label{fig:aroc_weight_cost_boxplot}
\end{figure}

We adjust weight-cost in training the RBM with three different values. Since this exponentially increases the number of networks to train as we stack up RBMs, a brute-force search for an optimal setting of weight-costs becomes very time-consuming. For example, when we have three hidden layers, we should fine-tune 27 different instances of pretrained networks. From our experiments, however, we observed that the good results tend to be obtained when the bottom layer has a small weight-cost and upper layers have progressively greater weight-costs. In order to validate the observation, we plot the statistics of AUC for a given weight-cost at each layer in Figure \ref{fig:aroc_weight_cost_boxplot}. For example, the left-most boxplot is computed from all combinations of weight-costs when the weight-cost in the first-layer RBM (L1) is fixed to 0.001 (this includes 9 combinations of weight-cost for three hidden layers. We count them for all different target sparsity and max-pooling size). For the first layer, the AUC goes up when the weight-cost is smaller. However, the trend becomes weaker through the second layer (L2) and goes opposite for the third layer (L3); the best AUC in median is obtained when the weight-cost is 0.1 for the third layer, even though the difference it slight. This result implies that it is better to encourage ``maximum likelihood" for the first layer by having a small weight-cost and regulate it for upper layers by having progressively greater weight-costs. This is plausible when considering the level of abstraction in the DNN that goes from acoustic feature summaries to semantic words.

\begin{figure}[t]
\centering
   \includegraphics[trim= 2mm -3mm 0mm 7mm, scale=0.64]{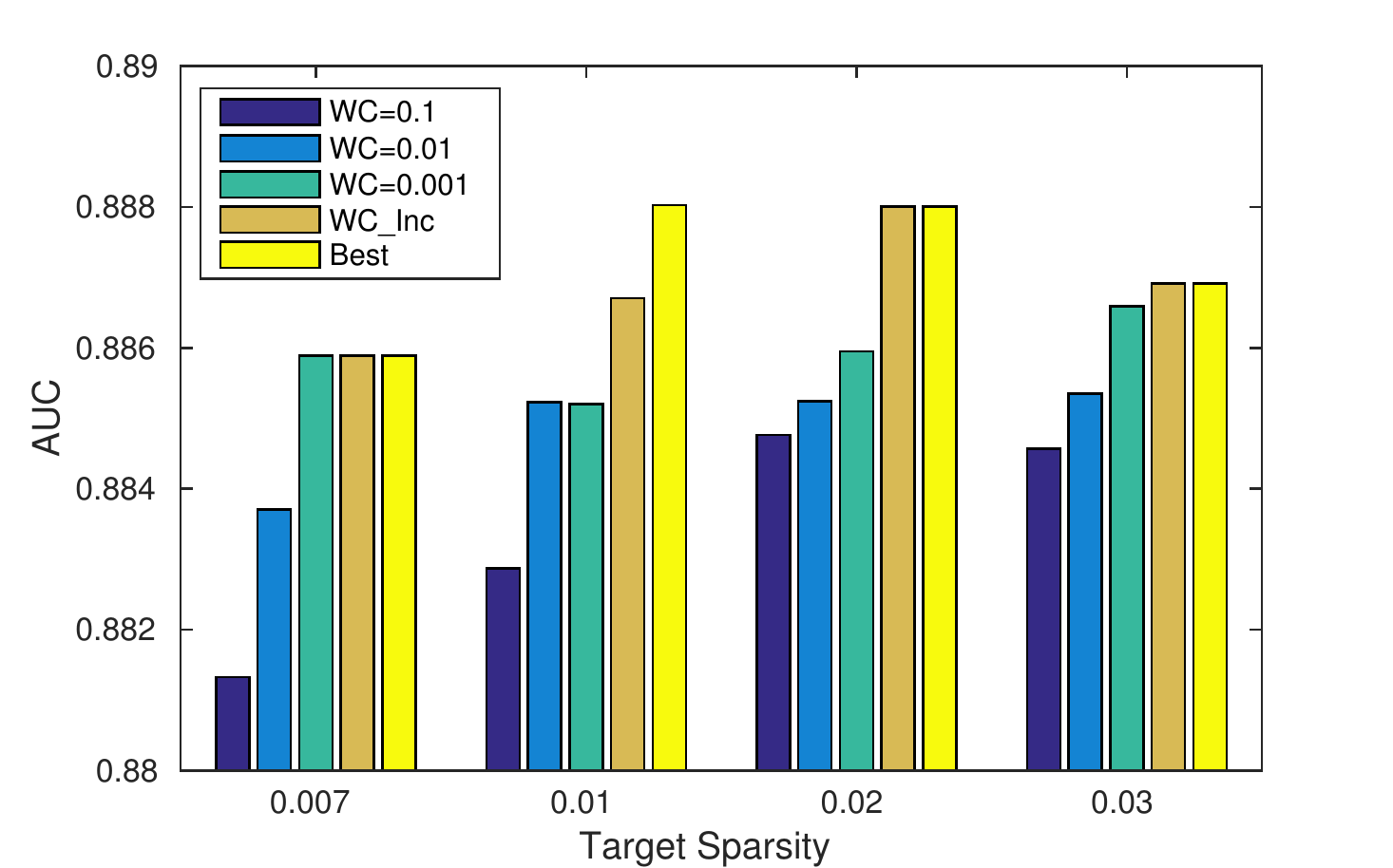}
 \caption{Results for different settings of weight-costs in training RBMs. ``Best" is the best result among all pretrained networks (27 instances). ``WC=0.1", ``WC=0.01" and ``WC=0.001" indicate when the weight-cost is fixed to the value for all hidden layers. ``WC\_Inc" means the best result among instances where the weight-cost is 0.001 for the bottom layer and it is greater than or equal to the value for upper layers (this includes 6 combinations of weight-costs for three hidden layers). The max-pooling size is fixed to 1 second here. }
  \label{fig:aroc_weight_cost}
\end{figure}

Based on this observation, we suggest a special condition for the weight-cost setting to reduce the number of pretraining instances. That is, we set the weight-cost to a small value (=0.001) for the first layer and an equal or increasing value for upper layers. Figure \ref{fig:aroc_weight_cost} compares the special condition denoted as ``WC\_Inc" to the best result and fixed settings for all layers. ``WC\_Inc" achieves the best result in three out of four and it always outperforms the three fixed setting. This shows that, with the special condition for the weight-cost setting, we can save significant amount of training time while achieving high performance.  



\subsection{Comparison with State-of-the-art Algorithms}
We lastly compare our proposed model to previous state-of-the-art algorithms in music auto-tagging. Since we use the MIREX 2009 version of Magnatagatune dataset for which Hamel et. al. achieved the best performance \cite{P.Hamel:11, P.Hamel:12},  we place their evaluation results only in Table \ref{table:results with prior arts}. They also used deep neural networks with a special preprocessing of mel-frequency spectrogram. However, our deep bag-of-features model outperforms them for all evaluation metrics. 


\section{Conclusion} \label{conclusion}
We presented a deep bag-of-feature model for music auto-tagging. The model learns a large dictionary of local feature bases on multiple frames selected by onset-based sampling and summarizes an audio track as a bag of learned audio features via max-pooling and averaging. Furthermore, it pre-trains and fine-tunes the DNN to predict the tags. The deep bag-of-feature model can be seen as a special case of deep convolutional neural networks as it has a convolution and pooling layer, where the local features are extracted and summarized, and has three fully connected layers. As future work, we will move on more general CNN models used in computer vision and train them with large-scale datasets.



\begin{table} [t]\renewcommand{\arraystretch}{1.4} \setlength{\tabcolsep}{3pt}
 \begin{center}
 \begin{tabular}{l|ccc cccc}
  \Xhline{2\arrayrulewidth}
  Methods & AUC-T & AUC-C & P3 & P6 & P9 & P12 & P15\\
  \Xhline{2\arrayrulewidth}
  PMSC+PFC \cite{P.Hamel:11} & 0.845 & 0.938 & 0.449 & 0.320 & 0.249 & 0.205 & 0.175\\
  PSMC+MTSL \cite{P.Hamel:11}  & 0.861 & 0.943 & 0.467 & 0.327 & 0.255 & 0.211 & 0.181\\
  Multi PMSCs \cite{P.Hamel:12}  & 0.870 & 0.949 & 0.481 & 0.339 & 0.263 & 0.216 & 0.184\\
  \bf{Deep-BoF} & \bf{0.888} & \bf{0.956} & \bf{0.511} & \bf{0.358} & \bf{0.275} & \bf{0.225} & \bf{0.190} \\
  \Xhline{2\arrayrulewidth}
 \end{tabular}
\end{center}
\caption{Performance comparison with Hamel et. al.'s results on the Magnatagatune dataset.}
\label{table:results with prior arts}
\end{table}

\section*{Acknowledgment}
This work was supported by Korea Advanced Institute of Science and Technology (Project No. G04140049).


\ifCLASSOPTIONcaptionsoff
  \newpage
\fi



\bibliographystyle{IEEEtran}
\bibliography{references}

\end{document}